# MiNet: A Convolutional Neural Network for Identifying and Categorising Minerals*

[1]E. B. Asiedu, [1]M. Agangiba and [1]D. Aikins
[1]University of Mines and Technology (UMaT), Tarkwa Ghana

## Abstract

Identification of minerals in the field is a task that is wrought with many challenges. Traditional approaches are prone to errors where there is no enough experience and expertise. Several existing techniques mainly make use of features of the minerals under a microscope and tend to favour a manual feature extraction pipeline. Deep learning methods can help overcome some of these hurdles and provide simple and effective ways to identify minerals. In this paper, we present an algorithm for identifying minerals from hand specimen images. Using a Convolutional Neural Network (CNN), we develop a single-label image classification model to identify and categorise seven classes of minerals. Experiments conducted using real-world datasets show that the model achieves an accuracy of 90.75%

## 1 Introduction

Minerals have been an essential part of society since the time of the prehistoric man. By definition, a mineral is a naturally occurring solid with definite chemical composition and an ordered internal structure (Rafferty, 2012; Bonewitz, 2012). They are mostly formed inorganically. Minerals and mineral resources are utilised in every section of society on a daily basis (King, 2019). They provide the basic needs – food, shelter, clothing and energy to mankind. The nutrients in food are supplied by minerals whiles various minerals are used in the building of shelters including aluminium, zinc, copper and lead. Coal or petroleum molecules are also used in the production of synthetic fibres.

Mineral resources are one of the essential natural resources that determine the economic and industrial development of a country. The mining sector is responsible for 60-90 per cent of Foreign Direct Investment (FDI) in low- and middle-income countries and represents 30-60 per cent of total exports in these countries (Anon., 2014). For instance, 45 per cent of government revenues in Botswana and 25 per cent in the Democratic Republic of Congo come from mining (Anon., 2014). In Ghana, the mining sector was responsible for 47.8 per cent of total exports in 2010 (McMahon and Moreira, 2014). The mining and quarrying sectors also contributed 15.8 per cent of state revenue in 2016 (Anon., 2016). Hence the impact of minerals on the development of an economy is immense; however, in order to realise the full benefits of minerals, there must be reliable ways of identifying them.

Identification of minerals is important to the process of defining their usage and forms a key part of the work of geologists. Rocks are aggregates in varying amounts of one or more minerals. However, certain contingencies would have to be in place in order for most minerals to form. Hence, by recognising the minerals in a rock, geologists can understand the origin of that rock (Egger, 2005). Understanding the composition of minerals also implies that geologists can use these insights to predict where significant economic deposits can be found (Egger, 2005). Attributes such as colour, shape, hardness, fracture and streak are used by geologists to discover the minerals that are present in a rock.

The precise identity of a mineral examined in the field is to some extent difficult to determine given that, the examination of minerals often takes place in distant locations (Rafferty, 2012). Geologists are typically not fitted to carry out thorough chemical and physical mineral analyses in stream beds and on mountainsides far from their laboratories. Rather, they conduct simple tests to deduce the properties of the mineral with which they can identify the mineral they are studying (Rafferty, 2012). Such experimental studies call for geologists to utilise experience and knowledge (Ault, 1998). In identifying a mineral, an observer should be capable of drawing a distinction between geologically germane and ungermane properties for the specimen under consideration and also understand that the properties are not certain for all samples (Ford, 2005). Consequently, identifying minerals is a task that requires a lot of skill and grit, which is not readily available to tyros in the field and non-geologists as well. Additionally, considering the sheer number of minerals, 4,507 as of January 2018 (Anthony *et al.,* 2018), even experienced mineralogists are fallible and are sometimes at their wits end in identifying some minerals.

Against this backdrop, this paper seeks to provide a simple and effective way of identifying and categorising minerals. The proposed solution is a deep learning model that is capable of distinguishing between hand specimen images of seven selected minerals. Deep learning is a sub-field of machine learning that focuses on learning consecutive layers of increasingly relevant representations (Chollet, 2017; Goodfellow *et al.*, 2016). These representations are learned through artificial neural networks, which are made up of layers that are arranged on top of each other. Learning in the context of deep learning means searching for a set of values for the parameters of the network so that the network will rightly map input examples to their related labels (Chollet 2017). Deep learning methods have greatly enhanced the state-of-the-art in computer vision, natural language processing and many other areas like pharmacology and biology (LeCun *et al.*, 2015). The most common deep learning approaches are Convolutional Neural Networks (CNNs) which are used for computer vision applications and Recurrent Neural Networks (RNNs) which are used for processing time-series data. RNNs have caused a quantum leap in the processing of sequential data such as text and speech, whereas CNNs have become the leading architecture for most image segmentation, classification, and detection tasks (LeCun *et al.*, 2015).

However, the application of CNNs to the task of identifying hand specimen images of minerals has largely been underexplored. Yao *et al.*, (2017) used principal component analysis (PCA) and Maximum Likelihood (ML) classification to identify rocks and their quartz content in the Gua Musang Goldfield. Izadi and Sadri, (2018) used thin sections in-plane and cross-polarized light mounted from rock samples to perform mineral segmentation and identification. Similarly, Thompson *et al.*, (2001), used 27 colour and texture parameters to train a three-layer feed-forward network using generalized delta error correction. In another study, Baykan and Yılmaz (2011), extracted six colour values of the pixels under the plane and cross-polarized light to train a multi-layer perceptron. These approaches utilised manually designed feature extraction pipelines coupled with a fully connected neural network other than a CNN. The main drawback of this approach is that the accuracy of the model is highly dependent on the design of the feature extraction pipeline which usually proves to be a difficult task (LeCun *et al.,* 1998). Zhang *et al.*, (2019) classified rock-mineral microscopic images using a pre-trained Inception-v3 model to perform transfer learning. Cheng and Guo, (2017) also used CNN to perform granularity analysis on thin section images. Although the work of Zhang *et al.*, (2019) and Cheng and Guo, (2017) are based on CNNs, however, none of the aforementioned papers dealt with the identification of hand specimen images.

This study, therefore, utilises a CNN to develop a single-label image classification model that identifies and categorises minerals from hand specimen images. CNNs require fewer parameters and are able to learn image-specific features, which makes them more suited for image classification tasks (O'shea and Nash, 2015). The proposed model, named MiNet can identify and categorise seven minerals namely: Biotite, Bornite, Chrysocolla, Malachite, Muscovite, Pyrite and Quartz. Further, this study opens up a traditionally underexplored area where intellectually interesting and challenging tasks and data sets can be discovered.

## 2 Resources and Methods Used

### 2.1 Data Collection

To obtain an appropriate dataset for this work, a web crawler was used to collect over 1200 images of the chosen minerals from across the internet. Inappropriate images (those that were not hand specimen pictures of minerals), as well as those that were wrongly formatted, were deleted. This reduced the dataset size to a total of 954 hand specimen pictures belonging to seven classes of minerals. The distribution of the images per class is shown in Fig. 1.

### 2.2 Data Pre-processing and Augmentation

Data pre-processing is done to prepare the raw data for further processing. Mean subtraction and normalisation were applied to each input data matrix. Mean subtraction involves subtracting the mean across every individual feature in the data and has the effect of zero-centring the data. Normalisation divides each zero-centred dimension by its standard deviation. This makes convergence faster while training the network. This is computed as shown in Equation (1):

$$X' = \frac{X - \bar{x}}{\sigma x} \#(1)$$

where $X'$ is the standardised dataset, $X$ is the original dataset, $\bar{x}$ is the mean of $X$ and $\sigma x$ is the standard deviation of $X$.

The images were resized to 256x256 and random crops of 224x224 were taken before being fed into the network. Additionally, these transformations were randomly applied to the images in the training set: horizontal and vertical flips, lightning and contrast changes, rotation, zoom and symmetric warp.

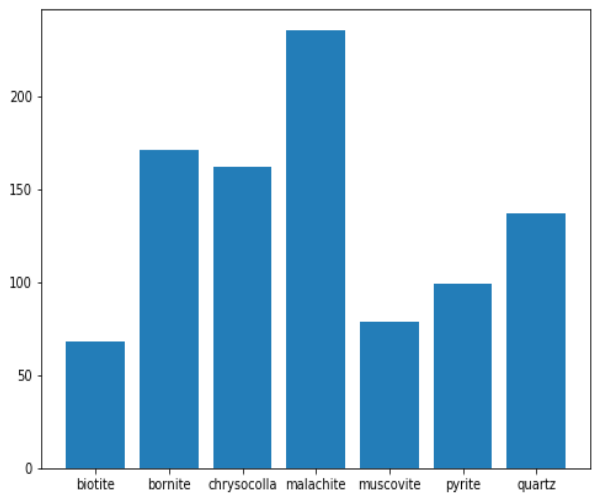

**Fig. 1 Distribution of images in the dataset**

### 2.3 Cross Validation

Cross validation is frequently used to evaluate machine learning models on small datasets (Jo *et al.*, 2020). MiNet was trained using *k*-fold cross validation as it guarantees that each sample is used for validation and also reduces pessimistic bias by using more training data (Raschka, 2018). In *k*-fold cross validation, we iterate over the dataset *k*-times and at each instance, the dataset is divided into *k* groups of approximately equal sizes. One fold is treated as a validation set and the model is trained on the remaining *k*-1 folds. The model's performance is computed as the average over the *k* performance estimates from the validation sets. In this paper, we use *5*-fold cross validation, which has been shown empirically to yield better test error rates (James *et al.*, 2013).

### 2.4 Model Architecture

Building an efficient neural network model calls for a thoughtful examination of the network architecture. The model proposed which is shown in Table 1 is a 121-layer Dense Convolutional Network (DenseNet) (Huang *et al.,* 2017).

DenseNets have fewer parameters as there is no need to relearn redundant feature-maps. They alleviate the vanishing gradient problem by improving the flow of information and gradients through the network. They have also been observed to have a regularising effect, which reduces overfitting on tasks with smaller training set sizes (Huang *et al.*, 2017). A softmax non-linearity is used on the final layer to compute the final class scores. These scores are the neural network's estimated probabilities for each class. The $i^{th}$ class probability $f(x)_i$ is computed as shown in Equation (2):

$$f(x)_i = \frac{e^{x_i}}{\sum e_j^x} \#(2)$$

The weights of a model pre-trained on ImageNet are used to initialise the network's weights (Deng *et al.*, 2009). The knowledge learned by the network on this task is transferred to the task of identifying minerals. This allows the model to achieve state-of-the-art results while still using a relatively small dataset. The model inputs a hand specimen image of a mineral and outputs the class probabilities for the input image.

### 2.4 Training

2.4.1 Loss Function

CNNs seek to optimise some objective function, specifically the loss function. This study utilised the standard categorical cross-entropy loss which is computed from the outputs of the final softmax layer. When the softmax activation function is combined with the cross-entropy loss, they form the softmax loss. The softmax loss can be written for the $i^{th}$ input feature $x_i$ with a label $y_i$ as shown in Equation (3):

$$L = \frac{1}{N}\sum_i L_i = \frac{1}{N}\sum_i -log\left(\frac{e^{f_{yi}}}{\sum_j e^{f_j}}\right) \#(3)$$

where N is the amount of training data and the $j^{th}$ element ($j \in [1, K], K$ is the number of classes) of the vector of class scores $f$ is represented by $f_j$.

2.4.2 Optimiser

MiNet was trained using AdamW (Loshchilov and Hutter, 2019) which addresses the problem of poor generalisation with the popular Adam optimisation algorithm (Kingma and Ba, 2014). Among other things, Loshchilov and Hutter (2019) point out that:

i. L$_2$ regularisation and weight decay are not identical. Although the two techniques can be made equivalent to SGD, this is not the case for Adam.

L$_2$ regularisation is not effective in Adam whereas weight decay is equally effective in both SGD and Adam

AdamW improves regularisation in Adam by decoupling the weight decay from the gradient-based update. This results in a substantial improvement in Adam's generalisation performance.

**Table 1 Architecture of MiNet**

| Layers | Output Size | MiNet |
|---|---|---|
| Convolution | $112 \times 112$ | $7 \times 7\ conv, stride\ 2$ |
| Pooling | $56 \times 56$ | $3 \times 3\ maxpool, stride\ 2$ |
| Dense Block (1) | $56 \times 56$ | $\begin{bmatrix} 1 \times 1 conv \\ 3 \times 3 conv \end{bmatrix} \times 6$ |
| Transition Layer (1) | $56 \times 56$ | $1 \times 1 conv$ |
| | $28 \times 28$ | $2 \times 2\ averagepool, stride\ 2$ |
| Dense Block (2) | $28 \times 28$ | $\begin{bmatrix} 1 \times 1 conv \\ 3 \times 3 conv \end{bmatrix} \times 12$ |
| Transition Layer (2) | $28 \times 28$ | $1 \times 1 conv$ |
| | $14 \times 14$ | $2 \times 2\ averagepool, stride\ 2$ |
| Dense Block (3) | $14 \times 14$ | $\begin{bmatrix} 1 \times 1 conv \\ 3 \times 3 conv \end{bmatrix} \times 24$ |
| Transition Layer (3) | $14 \times 14$ | $1 \times 1 conv$ |
| | $7 \times 7$ | $2 \times 2\ averagepool, stride\ 2$ |
| Dense Block (4) | $7 \times 7$ | $\begin{bmatrix} 1 \times 1 conv \\ 3 \times 3 conv \end{bmatrix} \times 16$ |
| Classification Layer | $1 \times 1$ | $7 \times 7\ globalaveragepool$ |
| | | $1000D\ fully-connected, softmax$ |

### 2.4.3 Learning Rate Finder and the One-Cycle-Learning Policy

The learning rate is a hyperparameter that controls how much the weights of a network are adjusted with respect to the loss gradient. Using the technique proposed by Smith (2018), a simple experiment is performed where the learning rate is gradually increased after each mini batch. The loss at each increment is recorded and plotted. Analysing the slope of the plot indicates the optimum learning rate. The best learning rate is associated with the steepest drop in the loss. A learning rate of $1e-2$ was chosen based on the result of the learning rate range test. The one-cycle-learning policy allows us to train a neural network easily. Following guidelines from Smith (2018), MiNet was trained using a cycle with two steps of equal lengths, one going from a lower learning rate to a higher one then back to the minimum.

Intuitively, it is helpful to oscillate the learning rate towards the higher learning rate. High learning rates prevent the model from landing in a steep area of the loss function, preferring to find a minimum that is flatter. The final part of the training with reducing learning rates allow the model to go inside the smoother part of a steeper local minimum.

## 3 Results and Discussion

The evaluation metric that is monitored in the experiments that were done in this study is the error rate which is the proportion of instances that have been incorrectly classified over the whole set of instances. The accuracy of the model can directly be inferred from the error rate as shown in Equation (4):

$$accuracy = 1 - error\ rate \quad \#(4)$$

The test results of the model (error rate, precision, recall) are shown in Table 3 and its confusion matrix is shown in Fig. 2. Inferring from equation 4 and Table 3, the accuracy of the model is 90.75%. From the confusion matrix of the test results (Fig. 2), the precision and recall for each class label have been computed and presented in Table 2. Precision is computed as shown in Equation (5):

$$precision = \frac{true\ positive}{true\ positive + false\ positive} \#(5)$$

The recall is also computed as shown in Equation (6):

$$recall = \frac{true\ positive}{true\ positive + false\ negative} \quad (6)$$

Precision quantifies the number of positive class predictions that actually belong to the positive class while recall quantifies the number of positive class predictions made out of all positive examples in the dataset.

**Table 2 Precision and Recall of the Individual Classes**

| Class | Precision | Recall |
|---|---|---|
| Biotite | 0.8462 | 0.8462 |
| Bornite | 0.9412 | 0.9143 |
| Chrysocolla | 0.9063 | 0.9355 |
| Malachite | 0.9565 | 0.9362 |
| Muscovite | 0.8125 | 0.9286 |
| Pyrite | 0.9 | 0.9 |
| Quartz | 0.963 | 0.9286 |

**Table 3 Test Results of the Model On All Classes**

| MiNet | |
|---|---|
| Error rate | 0.0925 |
| Average Precision | 0.9036 |
| Average Recall | 0.9127 |

For our task of classifying minerals, high precision is desirable because there is a high cost associated with making a wrong prediction. A wrong prediction would result in wrong deductions which could lead to a waste of time and resources in exploration activities. As such the proportion of positive predictions that are actually correct in our case, is more important than the proportion of actual positives that were predicted correctly. Although the model achieves a high precision of 0.9036, this is still lower than a recall of 0.9127. A look at the performance of the model on the individual classes shows that this is not a trend but that the relatively low precision can be attributed to two of the seven classes.

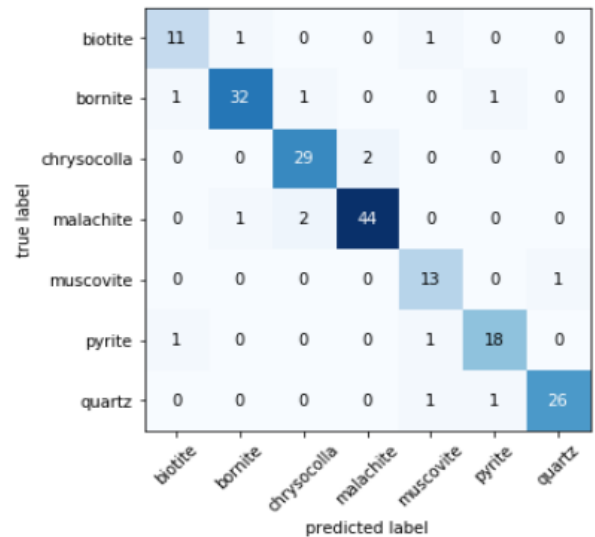

**Fig. 2 Confusion Matrix of Test Results**

The precision of the model is higher than the recall for three of the seven classes whereas they are equal in the case of biotite and pyrite. However, for muscovite, the recall is much higher than the precision. The precision and recall are 0.8125 and 0.9286 respectively. This may be attributed to the small number of muscovite samples in the dataset.

Despite the fact the model has higher precision than recall for most of the classes, it does this without sacrificing the completeness of the model. The recall of the model is still high even for the classes where the precision is higher. Two ResNet architectures with 18 and 34 convolutional layers were trained for the same number of epochs to compare their performance with the Dense Convolutional Network that was used in this work. Although both ResNet models have significantly more parameters than MiNet, their error rates are still higher than that of MiNet. This affirms the aforementioned parameter efficiency of DenseNets. Table 4 shows a comparison of the results of the two ResNet models with the results obtained with MiNet.

**Table 2 Comparison of MiNet with other Alternative CNN Architectures**

| Model | Number of Parameters | Error rate |
|---|---|---|
| ResNet 18 | 11,707,975 | 0.1114 |
| ResNet 34 | 21,816,135 | 0.0956 |
| **MiNet** | **8,011,655** | **0.0925** |

The study employed a one-cycle-learning policy during training to find a good minima of the loss function quickly. The learning rate and momentum were varied throughout one learning cycle, with the former increasing and then decreasing, and the latter varying inversely. To test the effectiveness and usefulness of this technique, another model was trained without it. This model achieved an error rate of 0.1512 which is higher than when the policy was utilised.

## 4 Conclusions and Recommendations

This study proposed MiNet – a single-label image classification model to identify and categorise seven classes of minerals. Using a DenseNet, the model achieved an accuracy of 90.75%. Although our method achieves compelling results on the dataset, the model does not consider all physical properties of the specimen in making its decisions. CNNs learn to detect visual features in an image, however, physical properties such as hardness and fracture cannot be perceived visually and as such, they are not leveraged by the model. The approach used in this paper can also be used to improve the performance of existing techniques for identifying minerals from microscopic images. MiNet would be a great tool to alleviate much of the stress involved in the identification of minerals in the field.

## Authors

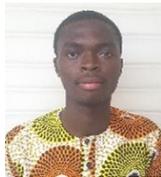
**E. Asiedu Brempong** is a Teaching Assistant at the Computer Science and Engineering Department of the University of Mines and Technology (UMaT), Tarkwa. He holds a Bachelor's degree in Computer Science and Engineering from UMaT. His research interests lie at the intersection of Machine Learning and Computer Vision and how these technologies can be leveraged to solve global challenges.

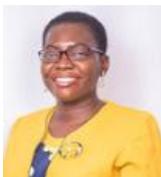
**M. Agangiba** is currently a lecturer at the Department of Computer Science and Engineering in University of Mines and Technology, Tarkwa. She obtained her PhD in Information Systems from the University of Cape Town (South Africa) and MSc from the Tver State Technical University (Russia) in Computer Complexes, Systems and Networks. She is a member of the Institute of Electrical and Electronics Engineering (IEEE), Association of Computing Machines (ACM), Association of Information Systems (AIS), Internet Society Chapter (ISOC), Ghana and the International Association of Engineers (IAENG). She is a Schlumberger Faculty for the Future Fellow and L'Oréal-UNESCO for Women in Science Fellow. Her research interest is in Web and Mobile Technologies, Interaction and Accessibility particularly in the areas of E- government and E-learning, Digital Inclusion for Persons with Disabilities, Information and Communication Technologies for Development.

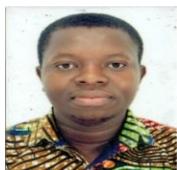
**D. Aikins** is a Lecturer at the University of Mines and Technology (UMaT), Tarkwa. He holds BSc and MPhil degree in Economic/Exploration Geology from the University of Mines and Technology (UMaT), Tarkwa. He researches in gold in the Birimian terrain of West Africa. He is a member of the International Association of Engineers (IAENG).